\documentclass[letterpaper]{article}
\usepackage{aaai}
\usepackage{times}
\usepackage{helvet}
\usepackage{courier}
\usepackage{color}
\usepackage{graphicx}

\begin{document}
%
\title{Cost-Effective HITs for Relative Similarity Comparisons}
\author{Michael J. Wilber\\Cornell University
\And Iljung S. Kwak\\University of California, San Diego
\And Serge J. Belongie\\Cornell University
}
\maketitle
\begin{abstract}
Similarity comparisons of the form ``Is object \emph{a} more similar to \emph{b} than to \emph{c}?'' are useful for computer vision and machine learning applications. Unfortunately, an embedding of $n$ points is specified by $n^3$ triplets, making collecting every triplet an expensive task.  In noticing this difficulty, other researchers have investigated more intelligent triplet sampling techniques, but they do not study their effectiveness or their potential drawbacks. Although it is important to reduce the number of collected triplets, it is also important to understand how best to display a triplet collection task to a user. In this work we explore an alternative display for collecting triplets and analyze the monetary cost and speed of the display. We propose best practices for creating cost effective human intelligence tasks for collecting triplets. We show that rather than changing the sampling algorithm, simple changes to the crowdsourcing UI can lead to much higher quality embeddings. We also provide a dataset as well as the labels collected from crowd workers.
\end{abstract}

\section{Introduction}
Recently in machine learning \cite{ckl,active_mds,tste,mcfee}, there has been a growing interest in collecting human similarity comparisons of the form ``Is \emph{a} more similar to \emph{b} than to \emph{c}?'' These comparisons are asking humans to provide constraints of the form $d(a,b) < d(a,c)$, where $d(x,y)$ represents some perceptual distance between $x$ and $y$. We will refer to these constraints as triplets. By collecting these triplets from humans, researchers can learn the structure of a variety of data sets. For example, the authors of \cite{mcfee} were able to learn music genres from triplet comparisons alone with no other annotations. Specifically in computer vision, human similarity comparisons are useful for creating perceptually-based embeddings. In \cite{agarwal2007generalized}, the authors created a two dimensional embedding where one axis represented the brightness of an object, and the other axis represented the glossiness of an object. In this work we focus on creating perceptual embeddings from images of food.

For any set of $n$ points, there are on the order of by $n^3$ unique triplets. Collecting such a large amount of triplets from crowd workers quickly becomes intractable for larger datasets. For this reason, a few research groups have proposed more intelligent sampling techniques \cite{ckl,active_mds}. However, the difficulty of collecting a large number of triplets is also related to the time and monetary cost of collecting data from humans. To investigate this relationship more closely, we chose to study a triplet human intelligence task (HIT). In this work we provide a better understanding of how the HIT design affects not only the time and cost of collecting triplets, but also the quality of the embedding, which is usually the researcher's primary concern.

\begin{figure}[tdp]
  \centering
\includegraphics[width=1\linewidth]{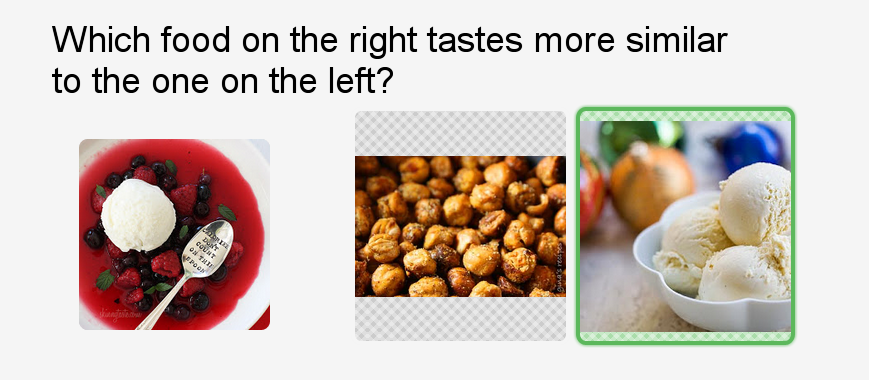}
\includegraphics[width=1\linewidth]{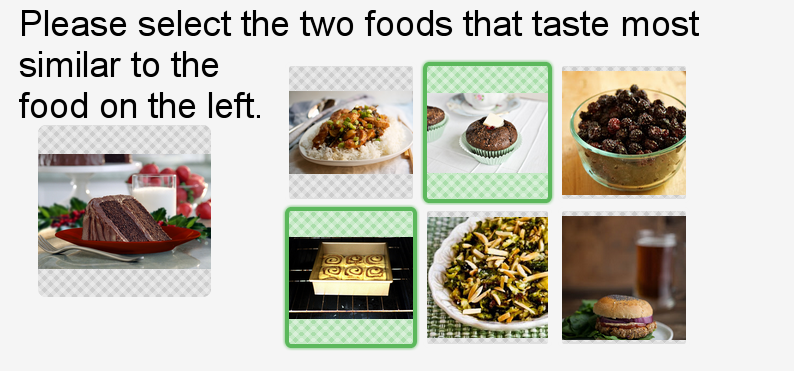}
\caption{\label{fig:uisample} Questions of the form ``Is object \emph{a} more similar to \emph{b} than to \emph{c}?'' have been shown to be a useful way of collecting similarity comparisons from crowd workers. Traditionally these comparsions, or triplets, would be collected with a UI shown at the top. In this work we collect triplets using a grid of images and ask the user to select the two most similar tasting foods to the food on the left. The grid UI, bottom, allows us to collect 8 triplets whereas the triplet UI, top, only yeilds a single triplet.}
\end{figure}


Traditionally, an MTurk task designed to collect triplets would show crowd workers three images, labeled \emph{a}, \emph{b}, \emph{c}. The worker is then asked to select either image \emph{b} or image \emph{c}, whichever looks more similar to image \emph{a}. See the top of Fig.~\ref{fig:uisample} for an example. Although this is the most direct design to collect triplets, it is potentially inefficient. Instead, we chose to investigate triplets collected from a grid of images. In the grid format, a probe image---analogous to image ``\emph{a}'' in the triplet representation---is shown next to a grid of $n$ images. The crowd worker is then asked to choose the $k$ most similar images from the grid. This layout allows us to collect $k$ images that are more similar to the probe image than the remaining $n-k$ images, yielding $k(n-k)$ triplets with one screen to the user. We can change the number of triplets per grid answer by varying $n$ and $k$, but this also affects the amount of effort a crowd worker must exert to answer the question. We are not the first to realize that a grid is more efficient for collecting triplets---such techniques were also used by \cite{catherine-cvpr-submission,ckl}---but we believe we are the first to investigate more thoroughly the effectiveness of triplets collected with a grid. This is important because previous authors acknowledge neither the efficiency gain nor the potential drawbacks of the grid triplets they rely on.


This paper outlines several UI modifications that allow researchers to multiply the number of triplets collected per screen for perceptual similarity learning. We show that simple changes to the crowdsourcing UI---\emph{instead} of fundamental changes to the algorithm --- can lead to much higher quality embeddings. In our case, using our grid format allows us to collect several triplet comparisons per screen. This leads to much faster convergence than asking one triplet question at a time. Researchers with tight deadlines can create reasonable embeddings with off-the-shelf algorithms and a low crowdsourcing budget by following our guidelines.

Our contributions are:
\begin{itemize}
\item A set of guidelines to use when collecting similarity embeddings, with insights on how to manage the trade-off between user burden, embedding quality, and cost;
\item A series of synthetic and human-powered experiments that prove our methods' effectiveness;
\item Evidence that each individual triplet sampled with a grid may capture less information than a uniformly random triplet, but that their quantity outweighs the potential quality decrease;
\item A dataset of 100 food images, ingredient annotations, and roughly 39\% of the triplet comparisons that describe it, to be made available upon publication.
\end{itemize}

\section{Related Work}




Perceptual similarity embeddings are useful for many tasks within the
field, such as metric learning~\cite{frome_2007}, image
search/exploration~\cite{geman}, learning semantic
clusters~\cite{crowdclustering}, and finding similar musical genres and
artists~\cite{tste,mcfee}. Our work is useful to authors who
wish to collect data to create such embeddings.
The common idea behind all of this work is that these authors use triplets to collect their embeddings.

In our work, we collect human similarity measurements of images in the form of triplets. The authors of \cite{crowdmedian} proposed an algorithm for collecting triplets from humans as well. However in \cite{crowdmedian}, the triplets that were collected did not have a probe image. because they formulated the question differently \cite{yi2013inferring} focuses on estimating user preferences from crowd sourced similarity comparisons. However \cite{yi2013inferring} uses pairwise comparisons rather than triplets.

\begin{figure}[h]
  \includegraphics[width=0.9\linewidth]{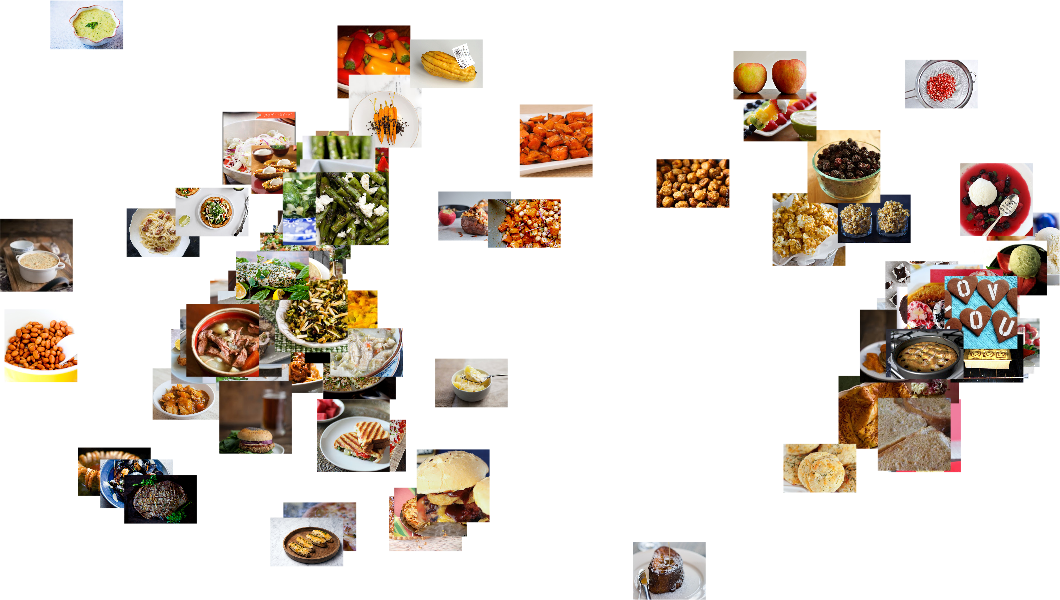}
  \rule{1\linewidth}{0.4pt}
  \includegraphics[width=0.8\linewidth]{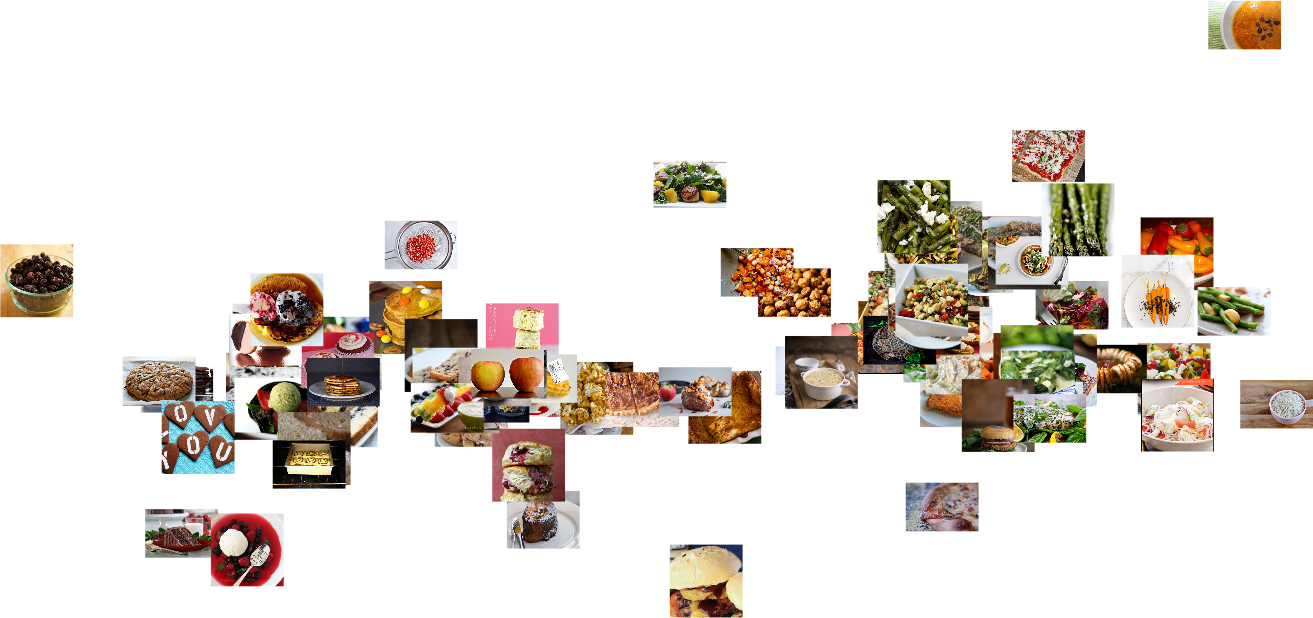}
  \caption{\label{fig:example-embedding} \textbf{Top:} An example
    cuisine embedding, collected with our 16-choose-4 grid UI
    strategy. This embedding cost us \$5.10 to collect and used 408
    screens, but yielded 19,199 triplets. It shows good clustering
    behavior with desserts gathered into the top left. The meats are
    close to each other, as are the salads.
\textbf{Bottom:} An
    embedding with 408 random triplets. This embedding also cost
    \$5.10 to collect, but the result is much dirtier, with worse
    separation and less structure. Salads are strewn about the right
    half of the embedding and a steak lies within the dessert area.
    From our experiments, we know that an embedding of such low
    quality would have cost us less than \$0.10 to collect using our grid
    strategy. }
\end{figure}

Our work bears much similarity to Crowd Kernel Learning~\cite{ckl} and Active
MDS~\cite{active_mds}. These algorithms focus on collecting triplets one at a time, but
sampling the \emph{best} triplets first. The idea behind these systems is
that the bulk of the information in the embedding can be
captured within a very small number of triplets, since most triplets convey redundant information. For instance, Crowd
Kernel Learning~\cite{ckl} considers each triplet individually,
modeling the information gain learned from that triplet as a
probability distribution over embedding space.
Active~MDS~\cite{active_mds} consider a set of triplets as a partial
ranking with respect to each object in the embedding, placing
geometric constraints on the locations where each point may lie. In our
work we focus on altering UI design to improve speed and quality of triplet collection.

\section{Method}

\begin{figure}[tdp]
\includegraphics[width=\linewidth]{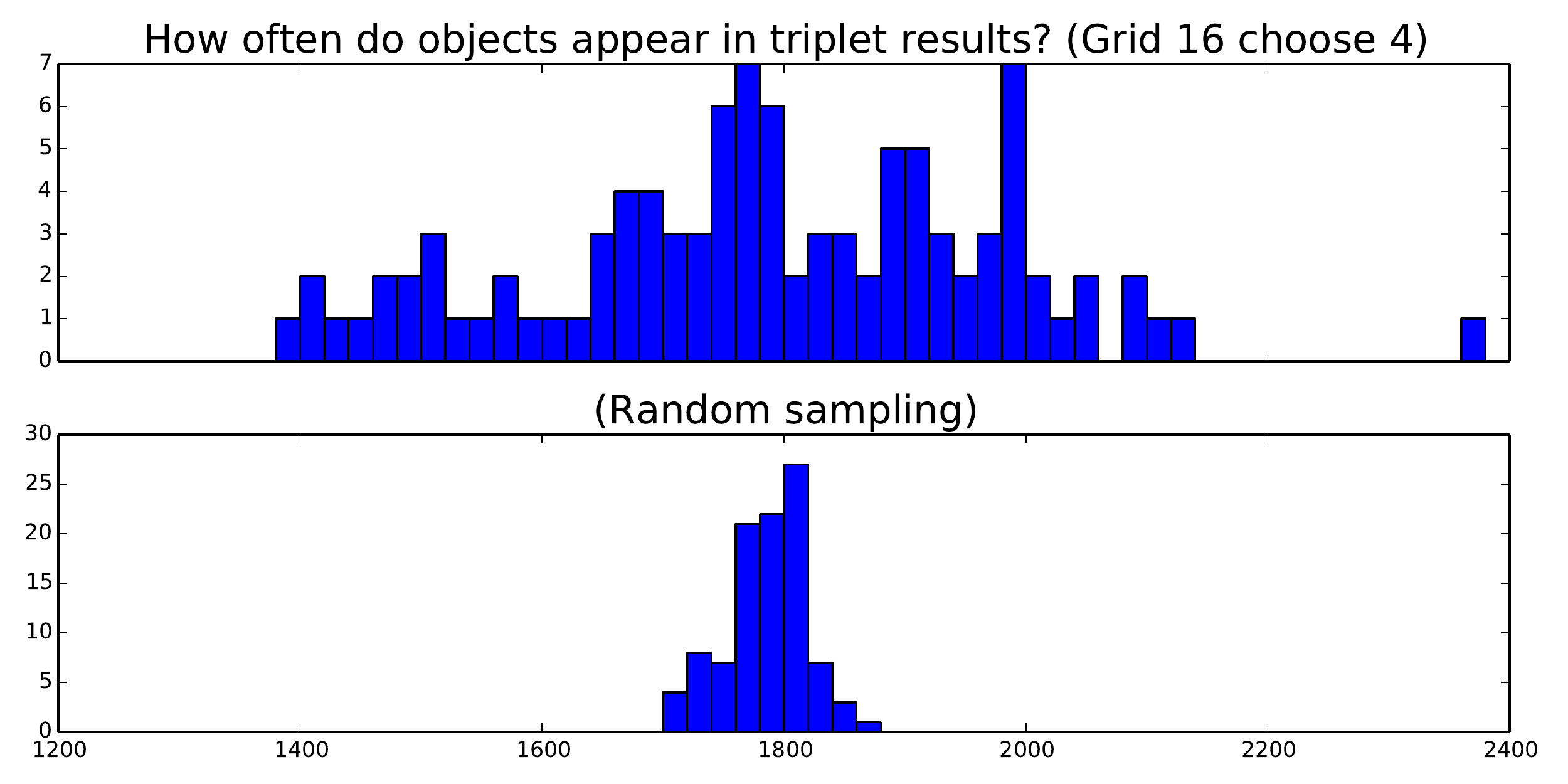}
\caption{\label{fig:distribution} \textbf{Random triplets have a
    different distribution than grid triplets.} The top histogram shows
  the occurrences of each object within human answers for ``Grid 16
  choose 4'' triplets. The bottom histogram shows a histogram of sampling
  random triplets individually. 59520 triplets were collected for both
  histograms. Each object occurs in our answers about $\hat{\mu}=1785$
  times, but the variation when using grid triplets (top) is much wider
  ($\hat{\sigma}\approx 187.0$) than the variation when sampling triplets
  uniformly (bottom, $\hat{\sigma}=35.5$). This effect is not recognized
  in the literature by authors who use grids to collect triplets. We study
  its impact in our experiments.}
\end{figure}
Instead of asking ``Is $a$ more similar to $b$ or $c$?'', we present
humans with a probe image and ask ``Mark $k$ images that are
most similar to the probe,'' as in Fig.~\ref{fig:uisample}. This way, with a grid of size $n$, a human
can generate $k \cdot (n-k)$ triplets per task unit.  This kind of query allows researchers to collect more triplets with a single screen. It allows crowd workers to avoid having to wait for multiple screens to load, especially in cases where one or more of the images in the queried triplets do not change. This also allows crowd workers to benefit from the parallelism in the low-level human visual system~\cite{wolfe_guided_1994}. Since many of these observations involve human issues, we conclude that the right way of measuring embedding quality is with respect to \emph{human cost} rather than the number of triplets. This \emph{human cost} is related to the time it takes crowd workers to complete a task \emph{and} the pay rate of a completed task. Some authors~\cite{catherine-cvpr-submission,ckl} already incorporate these ideas into their work but do not quantify the improvement. Our goal is to formalize their intuitive notions into hard guidelines.

%


It is important to note that the \emph{distribution of grid triplets
  is not uniformly random}, even when the grid entries are selected
randomly and even with perfect answers. To our knowledge, no authors
that use grids acknowledge this potential bias even though it
deteriorates each triplet's quality, as we will show in our experiments.
Figure~\ref{fig:distribution} shows a histogram of how many times each
object occurs in our triplet answers. When using grid sampling, some
objects can occur far more often than others, suggesting that the
quality of certain objects' placement within the recovered embedding
may be better than others. The effect is less pronounced in random
triplets, where objects appear with roughly equal frequency. This observation is important to keep in mind because the unequal distribution
influences the result.

\begin{figure*}[t]
  \includegraphics[width=\linewidth]{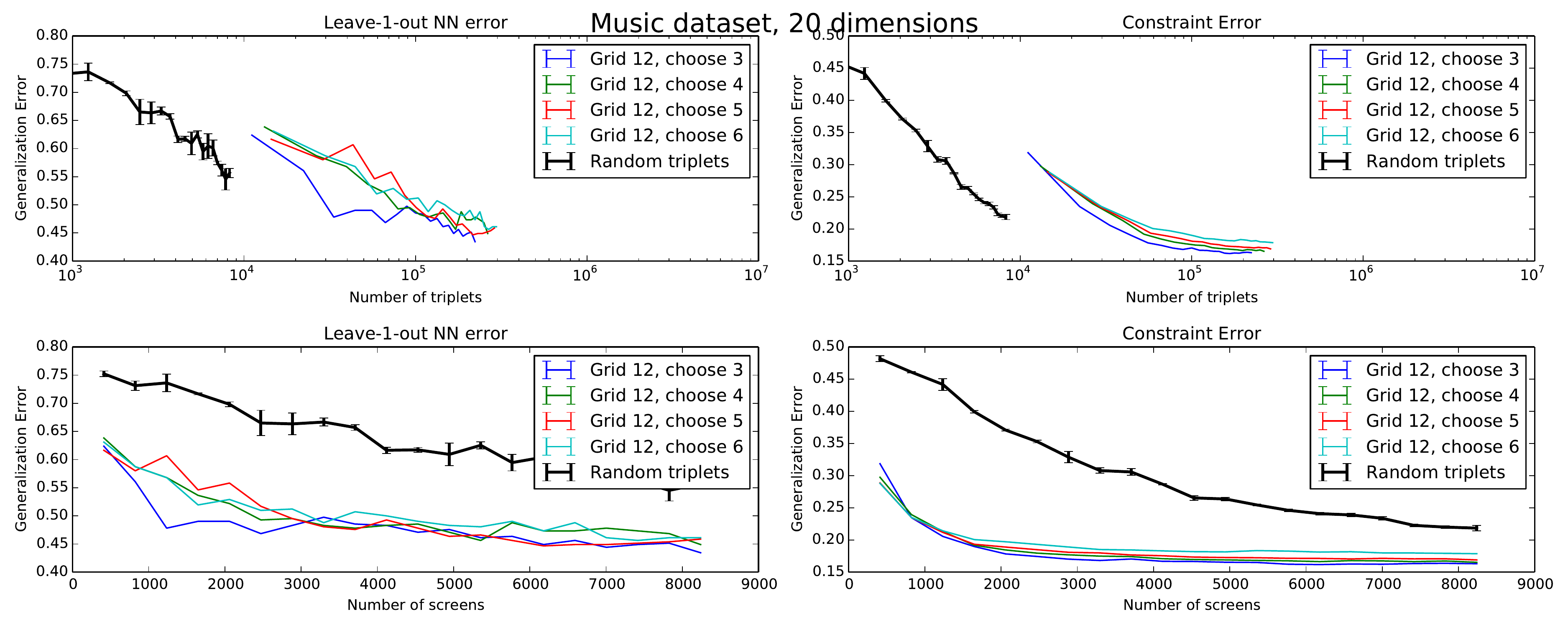}
  \caption{\label{fig:synthetic}Over the course of a synthetic
    experiment, we collect triplets, either randomly one at a time
    (thick black line) or in batches using our grid UI (colored
    lines). When the embedding quality is viewed as the number of
    triplets gathered (top two graphs), it appears that sampling
    random triplets one at a time yields a better embedding. However,
    when viewed as a function of human effort, grid triplets create
    embeddings that converge much faster than individually sampled
    triplets. Here, quantity outweighs quality as measured by
    Leave-One-Out NN Error (left graphs) and Triplet Generalization
    Error (right graphs). See text for details.}
\end{figure*}

\section{Synthetic Experiments}


We aimed to answer two questions: \emph{Are the triplets acquired from
  a grid of lower quality than triplets acquired one by one?} Second,
\emph{even if grid triplets are lower quality, does their quantity
  outweigh that effect?} To find out, we ran synthetic ``Mechanical
Turk-like'' experiments on synthetic workers. For each question, we
show a probe and a grid of $n$ objects. The synthetic workers use
Euclidean distance within a groundtruth embedding to choose $k$
grid choices that are most similar to the probe. As a baseline, we
randomly sample triplet comparisons from the groundtruth embedding
using the same Euclidean distance metric. After collecting the test
triplets,
we build a query embedding with t-STE~\cite{tste} and compare this
embedding to the groundtruth. This way, we can measure the quality of our embedding
with respect to the total amount of human effort, which is the number
of worker tasks. This is not a perfect proxy for human behavior, but
it does let us validate our approach, and should be considered in
conjunction with the actual human experiments that are described later.

\textbf{Datasets.} We evaluated our UI paradigm on three datasets.
First, we used MNIST1k, a handwritten digit dataset containing 1,000
random digits across 10 classes. To generate groundtruth comparison
triplets, we use Euclidean distance between feature vectors.
Second, we use the music similarity dataset from~\cite{tste} as a
point of comparison. This set contains 9,107 human-collected triplets
for 412 artists. Finally, we present results on a subset of
LFW~\cite{huang_labeled_2008}, the Labeled Faces in the Wild dataset.
We considered identities that have between 32 and 77 images in the
set, using the face attribute vectors extracted by~\cite{afs}. This
leaves us with a total of 938 73-dimensional feature vectors from 20
identities. To generate groundtruth triplets, we again considered
Euclidean distance. These three datasets provide us with a healthy
balance of synthetic and real-world nonvectorial data.

\textbf{Metrics.} Our goal is not to build a competitive face or
written digit recognizer; rather, we simply wish to evaluate the
quality of a perceptual embedding constructed with the help of
synthetic workers. To do this, we evaluate each embedding's quality
using two metrics from~\cite{tste}: Triplet Generalization Error,
which counts the fraction of the groundtruth embedding's triplet
constraints that are violated by the recovered embedding; and
Leave-One-Out Nearest Neighbor error, which measures the percentage of
points that share a category label with their closest neighbor within
the recovered embedding. As pointed out by~\cite{tste}, these metrics
measure different things: Triplet Generalization Error measures the
triplet generator UI's ability to generalize to unseen constraints,
while NN Leave-One-Out error reveals how well the embedding models the
(hidden) human perceptual similarity distance function. We use these
metrics to test the impact that different UIs have on embedding quality.






\textbf{Results.} Across all three datasets, our experiments show that
even though triplets acquired via the grid converge faster than random triplets, each individual grid triplet \emph{is of lower quality} than an individual random triplet.
Figure~\ref{fig:synthetic} shows how the music dataset embedding
quality converges with respect to the number of triplets. If triplets
are sampled one at a time (top two graphs), random triplets converge
much faster on both quality metrics than triplets acquired via grid
questions. However, this metric does not reveal the full story because
grid triplets can acquire several triplets at once. When viewed with
respect to the number of \emph{screens} (human task units), as in the
bottom two graphs in Figure~\ref{fig:synthetic}, we now see that the
grid triplets can converge far faster than random with respect to the
total amount of human work. This leads us to conclude that ``quality
of the embedding \emph{wrt.} number of triplets'' is the wrong metric
to optimize because framing the question in terms of triplets
  gives researchers the wrong idea about how fast their
  embeddings converge. A researcher who only considers the inferior
performance of grid triplets on the ``per-triplet'' metric will prefer
sampling triplets individually, but they could achieve much better
accuracy using grid sampling even in spite of the reduced quality of
each individual triplet, and as we shall see in our human experiments,
this translates into decreased cost for the researcher.
In other words, efficient collection UIs are better than random
sampling, even though each triplet gathered using such UIs does not
contain as much information.

Why does this happen? In all cases, the 12 images within the grid were
chosen randomly; intuitively, we expect a uniform distribution of
triplets. However, because certain objects are more likely than others
to be within each grid's ``Near'' set, certain objects will appear in
the triplet more often than others. This leads to a nonuniform
distribution of correct triplets, as shown in
Fig.~\ref{fig:distribution}. Here, we can see that the non-uniformity
creates a difference in performance.

The other two datasets---MNIST and Face---show very similar results so we do not report them here.
In all cases, any size of grid UI outperforms random selection.
However, we do see a small spread of quality across different grid
sizes. As in the music dataset, the error is lowest when we force our
synthetic workers to select 3 close images out of 12 as opposed to
selecting the 4, 5, or 6 closest images. This difference is more
pronounced in the ``Leave-One-Out NN'' metric. This could be because
selecting the 3 closest images allows the metric to be more precise
about that image's location in the embedding since it is compared to
fewer neighbors. Our synthetic workers always give perfect answers; we
do not expect imperfect humans to reflect this effect.

%
%

\begin{figure}[h!]
  \includegraphics[width=\linewidth]{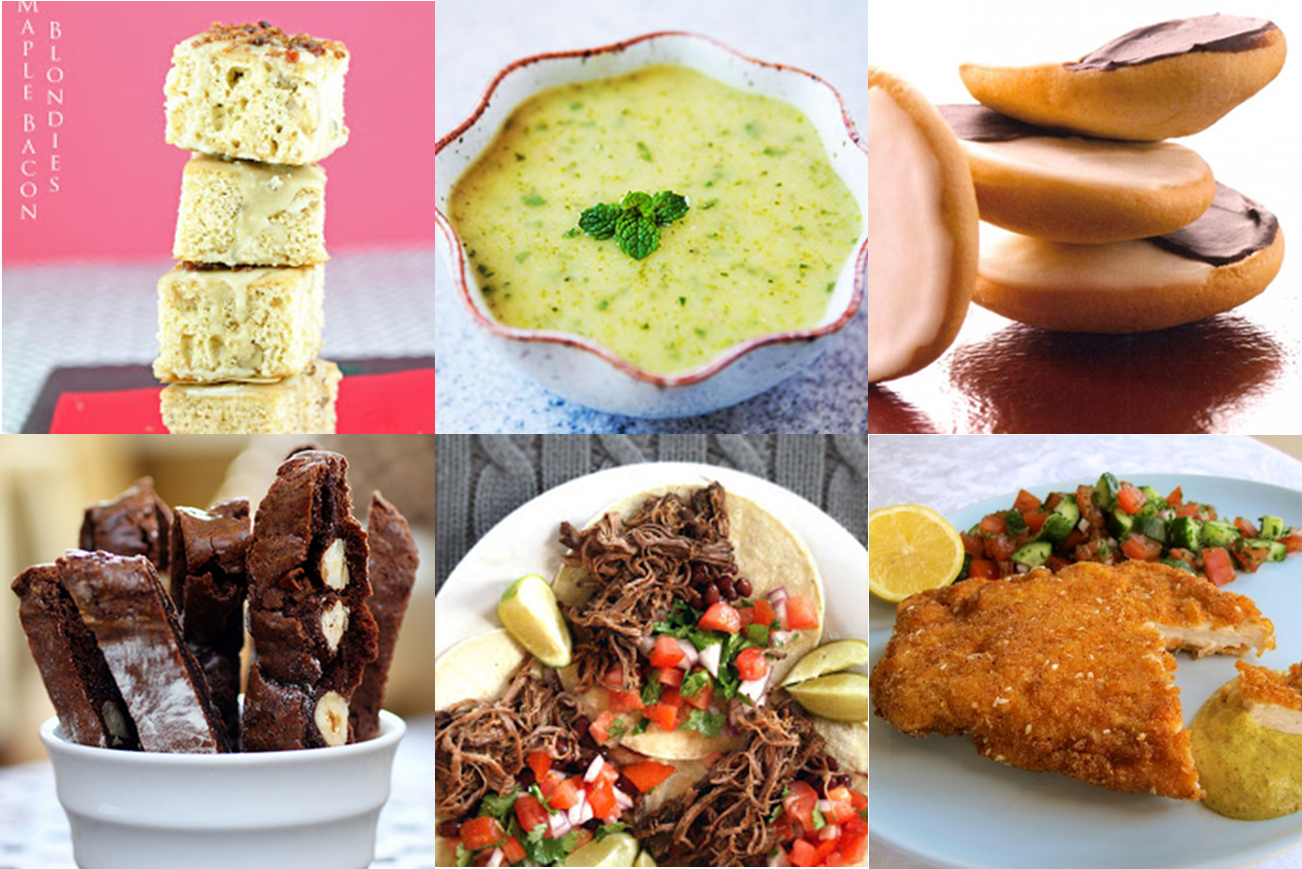}
  \caption{\label{fig:examplefoods} Example images from our dataset. The images
	in our dataset span a wide range of foods and imaging conditions. The dataset
	as well as the collected triplets will be made available upon publication.}
\end{figure}

\section{Human Experiments}

These synthetic experiments validate our approach, but they have
several problems. In particular, there is no reason why humans would
behave similarly to a proxy oracle as described above. Further, we
must also consider the effort of our workers, both in terms of the
time it takes to complete each task and how much money they can make
per hour---metrics that are impossible to gather via synthetic means.
To verify that these approaches build better embeddings even when humans
provide inconsistent triplets, we ran Mechanical Turk
experiments on a set of 100 food images sourced from Yummly recipes
with no groundtruth. The images were filtered so that each image contained
roughly one entree. For example, we avoided images of sandwiches with
soups. Example images are shown in Fig.~\ref{fig:examplefoods}. For each experiment, we allocated the same amount of money
for each hit, allowing us to quantify embedding quality with respect to cost.
Upon publication, the dataset as well as the collected triplets will be
available for download.

\textbf{Design.} For each task, we show a random probe and a
grid of $n$ random foods. We ask the user to select the $k$ objects
that ``taste most similar'' to the probe. We varied $n$ across $(4, 8,
12, 16)$ and varied $k$ across $(1,2,4)$. We ran three independent
repetitions of each experiment. We paid \$0.10 per HIT, which includes
8 usable grid screens and 2 catch trials. To evaluate the quality of the embedding returned by each
grid size, we use the same ``Triplet Generalization Error'' as in our
synthetic experiments: we gather all triplets from all grid sizes and
construct a reference embedding via t-STE. Then, to evaluate a set of
triplets, we construct a target embedding, and count how many of the
reference embedding's constraints are violated by the target
embedding. Varying the number of HITs shows how fast the embedding's
quality converges.

\begin{figure}[tdp]
  \includegraphics[width=\linewidth]{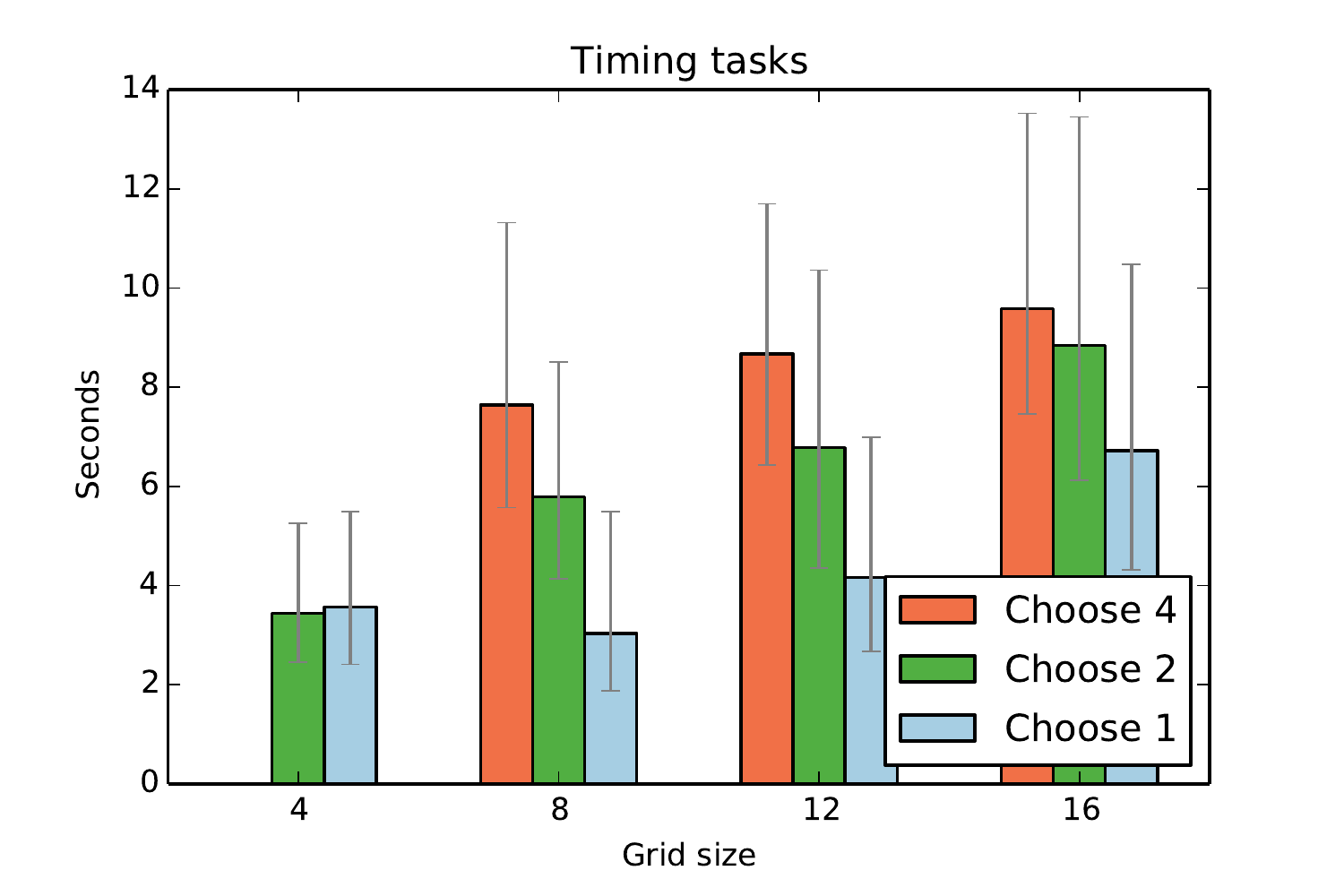}
  \caption{\label{fig:timing}  We
    show the median time that it takes a human to answer one grid. The
    time per each task increases with a higher grid size (more time
    spent looking at the results) and with a higher required number of
    near answers (which means more clicks per task). Error bars are 25
    and 75-percentile.}
\end{figure}

\begin{figure*}[t]
  \includegraphics[width=\textwidth]{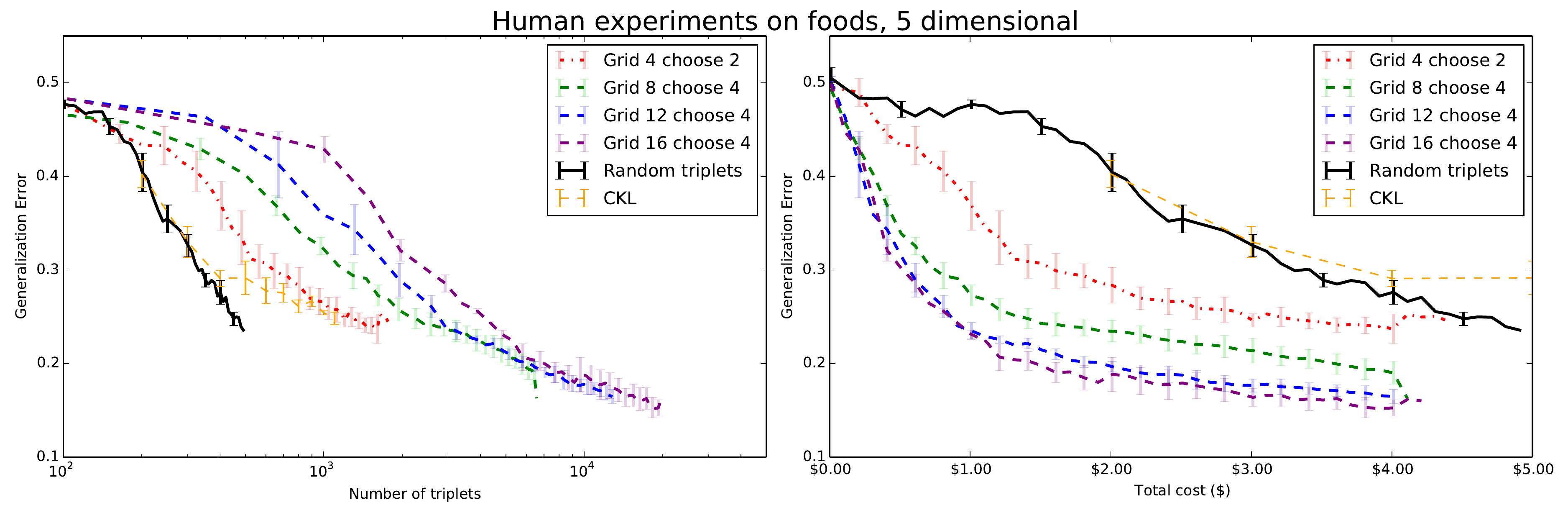}
  \caption{\label{fig:foodhuman} Results of our human experiments on
    the food dataset. Left graph: Triplet generalization error when
    viewed with respect to the total number of triplets. Right: The
    same metric when viewed with respect to the total cost (to us) of
    constructing each embedding. The left graph implies that a
    randomly-sampled embedding appears to converge faster. However,
    when quality is viewed with respect to cost, we find that an
    embedding generated using a 16-choose-4 grid cost \$0.75, while an
    embedding with random triplets of similar quality costs \$5.00. It
    is clear that the grid UI saves money; in this case, by over a
    factor of 6. }
\end{figure*}

\textbf{Baseline.} Since we wish to show that grid triplets produce
better-quality embeddings at the same cost as random triplets, we
should collect random $(a,b,c)$ comparisons from our crowd workers for
comparison. Unfortunately, collecting all comparisons one at a time is
infeasible (see our ``Cost'' results below), so instead, we construct
a groundtruth embedding from all grid triplets and uniformly sample
random constraints from the embedding. This is unlikely to lead to
much bias because we were able to collect 39\% of the possible unique
triplets, meaning that t-STE only has to generalize to constraints
that are likely to be redundant. All evaluations are performed
relative to this reference embedding.

\subsection{Results}
Two example embeddings are shown in Fig.~\ref{fig:example-embedding}.

\textbf{Cost.} Across all experiments, we collected 14,088 grids,
yielding 189,519 unique triplets. Collecting this data cost us
\$158.30, but sampling this many random triplets one at a time would
have cost us \$2,627.63, which is far outside our
budget\footnote{There are $100\cdot 99 \cdot 98 / 2 =485,100$
  possible unique triplets and each triplet answer would cost one
  cent. We additionally need to allocate 10\% to Amazon's cut and 20\%
  of our tasks are devoted to catch trials.}. If we had used the
16-choose-4 grid strategy (which yields 48 triplets per grid), we
would be able to sample all unique triplets for about \$140---a feat
that would cost us \$6737.50 by sampling one at a time.

\begin{table}[ht]
  \centering
  \small
\begin{tabular}{r|ccr}
Grid $n$ choose $k$&Error at \$1&Time/screen (s)&Wages (\$/hr)\\
\hline
$n$: 4,$\quad\;\,$ $k$: 1 & 0.468 & 3.57 & \textbf{\$10.09} \\
       $k$: 2 & 0.369 & 3.45 & \textbf{\$10.45} \\
$n$: 8,$\quad\;\,$ $k$: 1 & 0.400 & 3.04 & \textbf{\$11.85} \\
       $k$: 2 & 0.311 & 5.79 & \textbf{\$6.22} \\
       $k$: 4 & 0.273 & 7.65 & \$4.71 \\
$n$: 12,$\quad$ $k$: 1 & 0.406 & 4.17 & \textbf{\$8.64} \\
        $k$: 2 & 0.294 & 6.78 & \$5.31 \\
        $k$: 4 & 0.235 & 8.67 & \$4.15 \\
$n$: 16,$\quad$ $k$: 1 & 0.413 & 6.72 & \$5.36 \\
        $k$: 2 & 0.278 & 8.84 & \$4.07 \\
        $k$: 4 & 0.231 & 9.59 & \$3.76 \\
\hline
Random & 0.477 & -- & -- \\
\hline
CKL & 0.403 & -- & -- \\
\end{tabular}
\caption{\label{tab:results}
Results of our actual Mechanical Turk
  experiments. We ask workers to choose the $k$ most similar objects
  from a grid of $n$ images. We invest \$1 worth of questions, giving
  us 100 grid selections. When $n$ and $k$ are large, each
  answer yields more triplets.
  Large grids require more time to complete, but many of our tasks
  (bold) still pay a respectable wage of more than \$6 per hour.
}
\end{table}

\textbf{Quality.} As we spend more money, we collect more triplets,
allowing t-STE to do a better job generalizing to unseen redundant
constraints. All embeddings converge to lower error when given more
triplets, but this convergence is not monotonic because humans are
fallible and there is randomness in the embedding construction. See
Fig.~\ref{fig:foodhuman} for a graphical comparison of grids with size
4,8,12, and 16. When viewed with respect to the number of triplets,
random triplets again come out ahead; but when viewed with respect to
cost, the largest grid converges more quickly than others, and even
the smallest grid handily outperforms random triplet sampling.

This time, we observe a large separation between the performance of
various grid sizes. Grid 16-choose-4, which yields $4\cdot 12=48$
triplets per answer, uniformly outperforms the rest, with Grid
12-choose-4 (at $4\cdot 8 = 32$ triplets per answer) close behind.
Both of these outperform 8-choose-4 (16 triplets/answer) and
4-choose-2 (4 triplets/answer).

We also compare our performance with the adaptive triplet sampling
strategy of~\cite{ckl}. CKL picks triplets one-at-a-time but attempts
to select the best triplet possible to ask by maximizing the
information gain from each answer. In our experiments, it did not
outperform random sampling; further analysis will be
future work.

Though catch trials comprised 20\% of the grid answers we collected, we found that the results were generally of such high quality that no filtering or qualification was required.


\textbf{Time.} Fig.~\ref{fig:timing} shows how fast each human takes
to answer one grid question. Our smallest task was completed in 3.5
seconds ( ), but even our largest grid (16 choose 4) can be
completed in less than 10 seconds. Times varies widely between
workers: our fastest worker answered 800 questions in an average of
2.1 seconds per grid task for 8-choose-1 grids.




\textbf{Worker Satisfaction.}
At our standard 1${{\mathrm{c}\mkern-6.5mu{\mid}}}$-per-grid/\$0.10-per-HIT rate,
our workers are able to make a respectable income, shown in
Tab.~\ref{tab:results}. The smallest tasks net more than \$10/hour by
median, but even our largest task allows half of our workers to make
\$3.76 for every hour they spend. If the fastest, most skilled worker
sustained their average pace in 8-choose-1 grids, they could earn over
\$17 per hour.


Since there is a trade-off between grid size and worker income, it is
important to consider just how far we can push our workers without
stepping over the acceptable boundaries. Across all of our
experiments, we received no complaints, and our tasks were featured on
multiple HIT aggregators including Reddit's
\texttt{HitsWorthTurkingFor} subreddit and the ``TurkerNation'' forums
as examples of bountiful HITs. Our workers did not feel exploited.

According to the
\texttt{HitsWorthTurkingFor} FAQ~\footnote{http://reddit.com/r/HITsWorthTurkingFor/wiki/index}, ``the
general rule of thumb \ldots is a minimum of \$6/hour.'' Though HITs
below this amount may be completed, the best workers may pass for more
lucrative HITs. Being featured in forums such as
\texttt{HitsWorthTurkingFor} gave us an advantage since our hit was visible to a very large
audience of potential skilled turkers. Though
high payouts mean higher cost, in our case, the benefit outweighed the
drawback.





\section{Guidelines and conclusion}
Throughout this paper, we have shown 
that taking advantage of simple batch UI tricks can save researchers
significant amounts of money when gathering crowdsourced perceptual
similarity data. Our recommendations can be summarized as follows:
\begin{itemize}
\item Rather than collecting comparisons one-at-a-time, researchers
  \textbf{should use a grid} to sample comparisons in batch, or should
  use some other UI paradigm appropriate to their task. However,
  researchers should not assume that such ``batch'' comparisons are of
  identical quality to uniformly random sampling---this is a trade-off
  that should be considered.
\item If cost is an issue, researchers should \textbf{quantify their results
	with respect to dollars spent}. We found that using our simple UI paradigm can creates embeddings of higher quality than those created using algorithms that pick the best triplet one-at-a-time.
\item Researchers should \textbf{continuously monitor the human effort
    of their tasks,} so that they can calculate an appropriate target
  wage and stand a better chance of being featured on ``Good HIT''
  lists and be seen by more skilled Turkers.
\item When using grids to collect triplets, researchers should
  \textbf{consider the trade-off between size and effort.} Consider
  that an $n$-choose-$k$ grid can
  yield\begin{equation}k(n-k)\end{equation} triplets per answer. Since
  this has a global maximum at $n = 2k$, one appropriate strategy is
  to select the largest $n$ that yields a wage of \$6/hour and set $k$
  equal to $n/2$.
\end{itemize}

There are several opportunities for future work. First, we should
better quantify the relationship between $n$, $k$, and task completion
time to build a more accurate model of human performance. Second, we
should continue investigating triplet sampling algorithms such as
``CKL'' as there may be opportunities to adaptively select grids to
converge faster than random, giving us advantages of both strategies.

\section{Acknowledgments}
We especially thank Jan Jake\v{s}, Tomas Matera, and Edward Cheng for
their software tools that helped us collect grid triplets so quickly.
We also thank Vicente Malave for helpful discussions. This work was
partially supported by an NSF Graduate Research Fellowship award (NSF
DGE-1144153, Author~1) and a Google Focused Research award (Author~3).









{\footnotesize
\bibliographystyle{aaai}
\bibliography{sources}
}

\end{document}